\documentclass[10pt,twocolumn,letterpaper]{article}

\usepackage{iccv}
\usepackage{times}
\usepackage{epsfig}
\usepackage{graphicx}
\usepackage{amsmath}
\usepackage{amssymb}

\usepackage[utf8]{inputenc}
\usepackage{amssymb}
\usepackage{amsmath}
\usepackage{amsfonts}   
\usepackage{dsfont}
\usepackage{algorithm}
\usepackage{graphicx}
\usepackage[T1]{fontenc}       
\usepackage{url}            
\usepackage{booktabs}       
\usepackage{nicefrac}       
\usepackage{microtype}      
\usepackage{lipsum}
\usepackage{algpseudocode}
\usepackage{xcolor}
\usepackage{multirow}
\usepackage{comment}
\usepackage{booktabs}

\usepackage[breaklinks=true,bookmarks=false]{hyperref}

\iccvfinalcopy 

\def\iccvPaperID{11969} 

\ificcvfinal\fi

\begin{document}

\title{A Modular End-to-End Multimodal Learning Method for Structured and Unstructured Data}

\author{Marco D'Alessandro \hspace{0.5cm} Enrique Calabrés \hspace{0.5cm} Mikel Elkano
\\
\\
Neuraptic AI 
\\
{\tt\small \{marco.dalessandro, enrique.hernandez, elkano\}@neuraptic.ai}
}

\maketitle
\ificcvfinal\thispagestyle{empty}\fi

\begin{abstract}
Multimodal learning is a rapidly growing research field that has revolutionized multitasking and generative modeling in AI. While much of the research has focused on dealing with unstructured data (e.g., language, images, audio, or video), structured data (e.g., tabular data, time series, or signals) has received less attention. However, many industry-relevant use cases involve or can be benefited from both types of data. In this work, we propose a modular, end-to-end multimodal learning method called MAGNUM, which can natively handle both structured and unstructured data. MAGNUM is flexible enough to employ any specialized unimodal module to extract, compress, and fuse information from all available modalities.
\let\thefootnote\relax\footnotetext{$^1$Code available on GitHub: \url{https://github.com/neuraptic/magnum}}
\end{abstract}

\section{Introduction}

Multimodal AI has demonstrated remarkable proficiency in the representation learning of unstructured data, particularly focusing on vision, language, video, and audio modalities \cite{radford2021learning, rombach2022high,ho2022imagen,schneider2023archisound}. Various pre-training pipelines have been proposed to align different modalities for performing inherently multimodal tasks like Visual Question Answering, Image Captioning, Text-To-Image search, and generation \cite{zhang2024vision,bie2023renaissance,zong2023self}. Conversely, structured data, including tabular or time-series data, remains unexplored, despite being the most prevalent data type in real-world settings \cite{bughin2018notes}. Many industry-relevant scenarios illustrate the coexistence of structured and unstructured data, where the former serves as a complement, offering relevant information that would be inaccessible if solely relying on the latter. For example, in healthcare, patient records (structured data) combined with diagnostic images and doctor's notes (unstructured data) can enhance diagnosis accuracy and personalized treatment plans; in retail, product descriptions in natural language can be coupled with historical sales data for demand forecasting; and in finance, text reports on earnings, together with historical price and volume data, can be crucial for predicting asset prices. Nevertheless, harnessing the advantages of multimodal learning for both structured and unstructured data presents multiple challenges as the number of modalities, input sizes, and data heterogeneity grows. 

The majority of multimodal models exploit shared semantics between modalities, promoting joint pre-training in a shared semantic space \cite{bao2022vl, singh2022flava, radford2021learning}, yet their application is confined to unstructured data where there is a high correlation between modality-specific inputs (e.g. an image and its caption). On the other hand, only a very limited number of models try to achieve a joint representation of structured and unstructured data, but their application is restricted to tabular-language tasks \cite{yin2020tabert,deng2022turl,lin2020bridging}, and adapting them to different downstream tasks involves significant engineering (e.g. retrieving database entries using natural language queries). A more systematic and generalizable approach to multimodal representation learning for both structured and unstructured data is introduced by the LANISTR model \cite{ebrahimi2023lanistr}, which suggests joint pre-training for vision, language, and tabular modalities. However, as is customary with joint pre-training, foundational datasets are essential to enable the model to generalize to smaller trimodal datasets for specific downstream tasks, and such foundational datasets are almost impossible to find for many industrial domains. 

In this paper, we propose a modular, end-to-end multimodal learning method called MAGNUM$^1$ (Modality-AGNostic mUltimodal Modular architecture), which can natively handle both structured and unstructured data. MAGNUM is flexible enough to employ any specialized unimodal module to extract, compress, and fuse information from all available modalities. It can also leverage both modality-specific transfer learning and fine-tuning based on whether a pre-trained architecture is available for a certain modality or not.

The remainder of this paper is structured as follows: Section 2 discusses related work regarding our methodological approach. Section 3 describes the model architecture, training objectives, and fine-tuning processes. Sections 4 and 5 detail our experimental results and conclude with final thoughts, respectively.

\section{Related Work}

Our approach to the structured-unstructured multimodal learning problem is related to two broad topics: parameter-efficient learning and graph neural networks. In this section, we introduce each of them separately.

\subsection{Parameter-Efficient Learning}

The most common way to adapt foundational large general-purpose pre-trained models to downstream tasks is to fine-tune all the model parameters, which results in high computational costs and memory usage, and the need to store several copies of the fine-tuned model for different tasks. A lightweight alternative came from the parameter-efficient learning literature that proposed to update only a small number of extra parameters while keeping backbone parameters frozen \cite{he2021towards, lester2021power}. Several methods have been proposed to flexibly adapt pre-trained backbones to different downstream tasks according to this logic. Adapter-tuning \cite{houlsby2019parameter, houlsby2020k} interleaves transformer layers with a feed-forward bottleneck module with skip-connection to adapt the layer's output before passing to the next layer. Prefix-tuning \cite{li2021prefix, wang2022dualprompt, jia_visual_2022} prepends tunable prefix vectors as learnable embeddings to the keys and values of the multi-head attention layers in transformers \cite{vaswani2017attention}. In prompt-tuning \cite{lester2021power, wang2022learning}, a set of learnable embeddings is prepended to the input embeddings from the first layer, and the augmented input is then normally processed by the frozen transformer layers.

\subsection{Graph Neural Networks}

Graph Neural Networks (GNNs) have recently emerged as a transformative paradigm in AI, due to the popularity of network-like interconnected data structures in various industrial and scientific domains. In a standard GNN, feature vectors are seen as nodes of a graph, and the message-passing reflects nodes collecting information from the neighborhood around it. Each node embedding has more data from distant reaches of the graph as message-passing iterations progress \cite{zhou2020graph}. GNN models have been proposed to account for a variety of problem solutions. In node classification tasks, node representations are learned to accurately predict node labels (e.g. node labels could be scientific topics in citation networks, or gender
and other attributes in social networks) by leveraging structural information of the networked data \cite{kipf2016semi, hamilton2017inductive}. Graph representation learning was used in biochemical domains for leveraging topological information in molecule systems for drug discovery and molecule classification \cite{wieder2020compact, wang2023graph, zhang2022graph}. Graph clustering techniques have been employed to find local inhomogeneous distributions of edges in many real-world systems \cite{chen2018harp, tsitsulin2023graph}, and for improving graph compression to represent intensive network structure with smaller sub-graphs \cite{ying2018hierarchical, diehl1905edge}.

\section{MAGNUM: A Modality-Agnostic Multimodal Modular Architecture}

In this section, we describe in detail the functional components of MAGNUM, and provide a description of the training objective and some considerations about the model fine-tuning.

\subsection{Model architecture}

\begin{figure*}
    \begin{center}
        \includegraphics[width=1\linewidth]{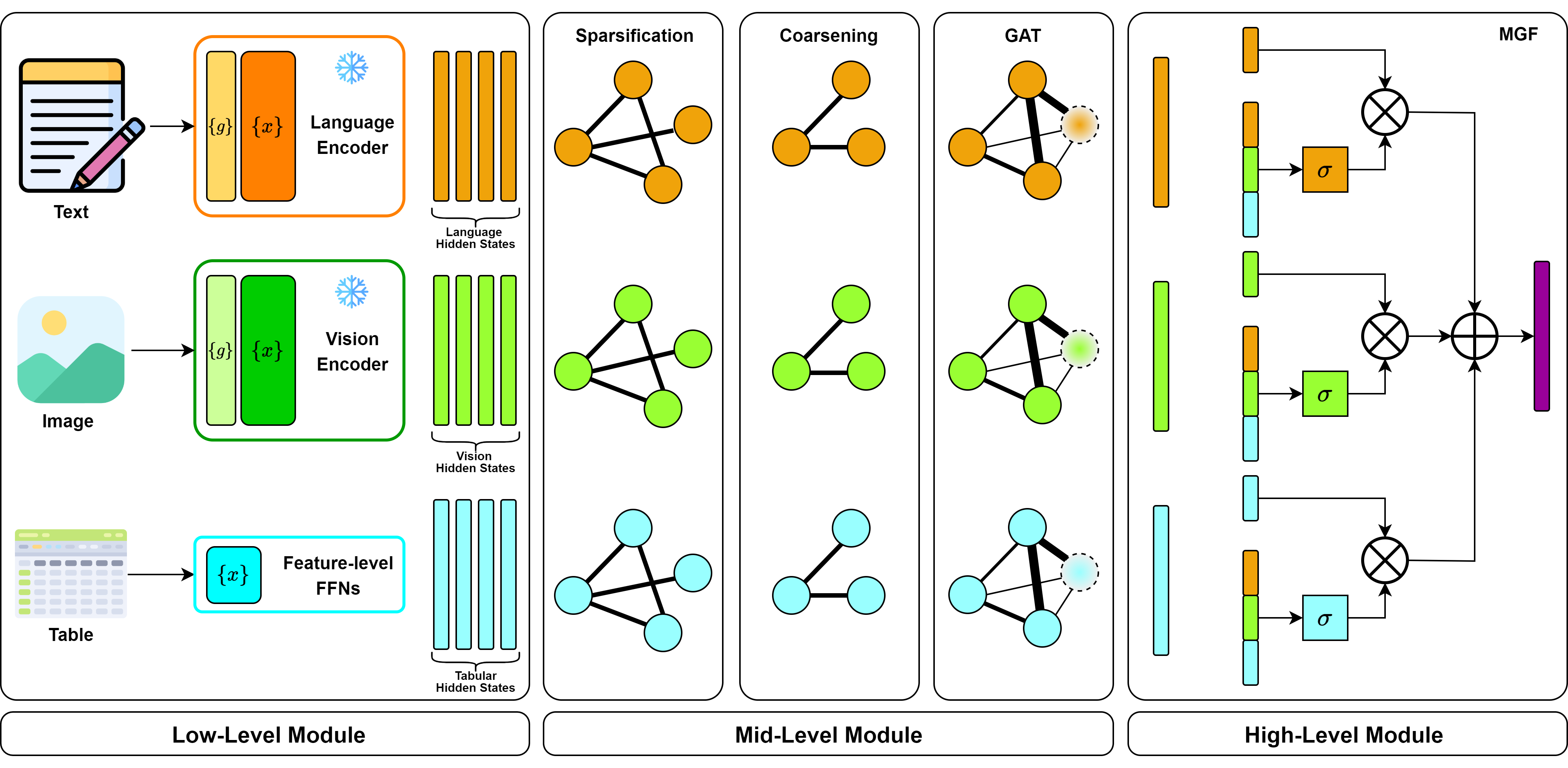}
    \end{center}
   \caption{A schematic view depicting the MAGNUM end-to-end pipeline. In the low-level module, unstructured data (i.e. text, image) are processed through transformer encoders, and structured data (i.e. tabular) through feature-level FFNs. In both cases, a set of hidden states is obtained for every modality. In the mid-level module, the hidden states go through three GNN-based steps in order to obtain a smaller set of hidden states. These are processed by a Multimodal Gated Fusion (MGF) layer in the high-level module. A final output hidden state is obtained through aggregation.}
    \label{fig:Fig.1}
\end{figure*}

Figure \ref{fig:Fig.1} summarizes the model architecture of MAGNUM, which is composed of three main modules: (1) low-level module for parameter-efficient feature extraction, (2) mid-level module for hidden representation compression, and (3) high-level module for multimodal fusion. \\

\noindent \textbf{Low-Level Module}. The purpose of this module is to extract features from several modalities and bring unimodal hidden states to the same dimensional space. We assumed two possible ways of deriving hidden states, one leveraging transfer learning in pre-trained architectures, and one generating hidden states from scratch when pre-training is not available or achievable for a specific modality (e.g. tabular). For simplicity, we refer to the case where the unstructured (U) modality is vision, or language, and the structured (S) modality is tabular, even though MAGNUM can handle any possible modality. 

In case pre-training is available, like the standard case of unstructured modalities, a frozen transformer encoder is used as features extractor with the aid of learnable prompts. Consider $f_i^U$ as the $i^{\textrm{th}}$ layer of the transformer encoder where $i=1,\ldots,K$ and $K$ is the total number of layers. Consider a set of $L$ learnable prompts, $\boldsymbol{g}=[g_1,\ldots,g_L]$, where $g_l \in \mathbb{R}^{L \times d^{\textrm{U}}}$, where $d^{\textrm{U}}$ is the embedding dimension of the transformer encoder for a given modality. The input embeddings now follow the form $[\boldsymbol{g}, \boldsymbol{x}] = [g_1,\ldots,g_L,x_{CLS},x_1,\ldots,x_J]$ where $x_j$ is the $j^{\textrm{th}}$ patch or token embedding in case of vision or language modality, respectively, $J$ the number of total embeddings, and $x_{CLS}$ is the class token embedding. The forward pass is described as follows:

\begin{equation}
    [\boldsymbol{g}_1, \boldsymbol{h}_1] = f_1^U([\boldsymbol{g}, \boldsymbol{x}])
    \hspace{1cm}
    i=1\
    \label{eq.1}
\end{equation}

\begin{equation}
    [\boldsymbol{g}_i, \boldsymbol{h}_i] = f_i^U([\boldsymbol{g}_{i-1}, \boldsymbol{h}_{i-1}])
    \hspace{1cm}
    i=2,3, \ldots, K
    \label{eq.2}
\end{equation}

\noindent where $[\cdot,\cdot]$ refers to vertical concatenation, $\boldsymbol{h}_i$ and $\boldsymbol{g}_i$ are hidden states related to the input embeddings and the prompts, respectively, and we omit the $U$ superscript for clarity of notation. A subset, $\boldsymbol{h}^U$, of hidden states of the last layer, K, is taken as the final hidden states representation of the given unstructured modality such that $\boldsymbol{h}^U = [\boldsymbol{g}_K, h^{(CLS)}_K]$, where $h^{(CLS)}_K$ is the hidden state related to the class token.

As mentioned above, structured modalities are managed differently when pre-trained encoders are not available. Consider a table consisting of the set of features $\boldsymbol{c} = [c_1,c_2,\ldots,c_N]$ where N is the number of columns in the dataset. We introduce a set of feed-forward neural networks (FFNs) for each feature projection $\boldsymbol{f}^S=[f^S_1,f^S_2,\ldots, f^S_N]$ such that $\Theta_{f_i} \in \mathbb{R}^{d^S}$, where $d^S$ is a desired embedding dimension for the given structure modality. The forward pass is described as follows:

\begin{equation}
    [h_1,h_2,\ldots,h_N] = [f^S_1(h_1), f^S_2(h_2), \ldots, f^S_N(h_N)]
    \label{eq.3}
\end{equation}

\noindent and the final hidden states representation of the given structured modality is then $\boldsymbol{h}^S=[h_1,\ldots,h_N]$.

In the final step of the low-level module, the sets of structured and unstructured hidden states are then projected to a common dimensional space via modality-specific FFN with common output dimension $d$.
\\

\noindent \textbf{Mid-Level Module}. Here, a stack of GNN-based processing steps turns modality-specific hidden states into a more dense and compressed set, which results in a smaller subset of hidden representations. The goal of this module is to obtain the best non-redundant and compressed representations from the unimodal hidden states extracted from the low-level module. The main principle of the current module is to consider hidden states as nodes of a graph, and apply graph-based computations to them. To do that, hidden states go through three steps: (1) \textit{sparsification}, (2) \textit{coarsening}, (3) \textit{attention}. \\

In \textit{sparsification}, an actual graph is drawn upon the raw hidden states via graph clustering \cite{donath1973lower, hall1970r}, such that the graph $G=(V,E)$, where $V = \boldsymbol{h}^m = [h_1,h_2,\ldots,h_I]$ is the set of vertices (or nodes), with $I$ being the number of hidden states for modality $m$ coming from the final low-level module projection, $h_i \in \mathbb{R}^d$, and $E$ is the set of edges to be estimated. Although diverse spectral efficient techniques have been proposed to account for the problem of graph clustering during the last year \cite{shi2000normalized, ng2001spectral, chan1992spectral}, we opted for a simple and fast kNN-based algorithm, which belongs to the family of kernel k-means methods \cite{dhillon2007weighted}. For every modality, the set of edges $E^m$ is then derived as well as its relative sparse adjacency matrix $A^m=|V^m| \times |V^m|$, where $m=1,\ldots,M$ and $M$ is the number of modalities processed by MAGNUM. \\

\noindent The modality-specific graph is further processed to obtain a compressed representation as a smaller graph via edge contraction \cite{lee2019self,diehl1905edge, ying2018hierarchical}. The \textit{coarsening} step is then meant to properly merge hidden states via edge pooling, where edges are defined in the previous step and stored in $A^m$. Here, we adopt the EdgePool algorithm \cite{diehl1905edge}, which solves the problem of edge pooling and node merging via a learnable pooling layer, which is then differentiable. The forward pass is as follows:

\begin{equation}
    [h^*_1, \ldots, h^*_T] = \textrm{EdgePool}([h_1, \ldots, h_N], A | \Theta)
    \hspace{1cm}
    T < N
    \label{eq.4}
\end{equation}

\noindent where we omit the superscript for modality $m$ for readability. As we can see, the output nodes, or hidden states, consist of a smaller set of nodes, where $h^*_t$ is a linear combination of adjacent nodes $(h_{n_i}, h_{n_j})$ based on provided $A$ and parameters $\Theta$. \\

\noindent The resulting set of nodes is finally passed through a graph attention layer, in the \textit{attention} step. Here, we adopt the dynamic graph attention variant proposed in \cite{brody2021attentive}. The hidden states enter as the only input to the graph attention model, and edge weights are estimated through neural networks. The goal is to obtain the final representations of the subset of hidden states obtained in the previous step. To do that, we place a learnable virtual node together with the hidden states set and assume a fully connected graph with no self-loops, so that all the nodes can share information between them and with the virtual node. After the application of the graph attention, the virtual node $h_m \in \mathbb{R}^d$ is used as the final representation for modality $m$.
\\

To sum up, the mid-level module first processes modality-specific raw hidden states coming from the bottom-level module and generates an adjacency matrix based on kNN node clustering. Next, it uses such an adjacency matrix for coarsening the graph into a smaller and compressed graph where the amount of nodes, or hidden states, is reduced. Finally, the hidden states are processed via graph attention, and the final hidden state representations are provided as output. 
\\

\noindent \textbf{High-Level Module}. In this module, the actual multimodal fusion takes place. The final unidimensional modality-specific representations coming from the previous module are combined in a unique embedding. We endow this module with a mechanism to filter out information coming from irrelevant modalities. By taking inspiration from the standard GRU and LSTM flow control in recurrent architectures and multimodal gated units \cite{arevalo2017gated}, we introduce Multimodal Gated Fusion (MGF) as modalities aggregation mechanism. The MGF takes as input a feature vector (the hidden states coming from the mid-level module) associated with modality $m$, $h_m$, $m=1,\ldots,M$, with $M$ being the total number of modalities considered. For each modality input, there is a gate neuron, $\sigma_m$, that controls the contribution of the modality-specific hidden state to the overall output. The modality-specific hidden state representation passes through an FFN and processed via a \textit{tanh} activation before being filtered by the gate. The modality-specific gate receives information from all the modalities involved, so that also modality $i$, $i \neq m$, contributes to properly encode relevant information about modality $m$. The representations obtained after the gating mechanism are then aggregated via summation. The forward pass is as follows:

\begin{equation}
    \begin{split}
        x_m & = \textrm{tanh}(f_m(h_m)) \hspace{1cm} m=1,2,\ldots,M \\
        \sigma_m & = \textrm{sig}(f^{\sigma}_m([h_m || h_i, \ldots, h_I])) \hspace{1cm} i \neq m \\
        z_m & = x_m * \sigma_m \hspace{1cm} m=1,2,\ldots,M \\
        h & = \sum_m z_m \hspace{1cm} m=1,2,\ldots,M
    \end{split}
\end{equation}

\noindent where $||$ is the horizontal concatenation operation, sig indicates the sigmoid activation function, and $f_m$ and $f^{\sigma}_m$ are modality-specific FFNs.

\subsection{Training objectives}

In this proposal, we are only dealing with the case where the final aggregated output from MAGNUM serves decision-making for classification problems. In this respect, the final aggregated hidden state can be used as a multimodal representation for any downstream task by simply applying a further FFN layer, $f^C$ on top of it. Given the final aggregated hidden state vector $h$, the training objective consists of minimizing the following quantity:

\begin{equation}
    \mathcal{L} = -\log [\mathbb{I}_y * \textrm{soft}(f^C(\textrm{MAGNUM}(x_1,\ldots,x_m)))]
\end{equation}

\noindent with MAGNUM being the stacking of low-, mid-, and high-level modules, $x_m$ the input for modality $m$, $m=1,\ldots,M$, \textit{soft} the softmax activation function, and $\mathbb{I}_y$ a one-hot encoded vector indicating the position related to the correct label.

\subsection{Fine-tuning MAGNUM}

The primary computational demand of MAGNUM stems from the forward passes of low-level modules, particularly when handling a greater volume of unstructured data or modalities. However, the adoption of prompt-tuning significantly lowers training expenses while maintaining the knowledge base of pre-trained encoders. Conversely, structured data necessitate only a feature-level conversion into a higher dimensional space. Within the MAGNUM framework, the encoders for unstructured modalities remain unchanged, with only the learnable prompts assimilating knowledge through the propagation of layers, serving as the conclusive output for a specific unstructured modality. This approach leads to early and substantial data compression, as all residual embeddings, whether they are patches for vision or tokens for language modalities, are excluded. MAGNUM is agnostic regarding the encoder models adopted for processing unstructured modalities provided they are pre-trained and compatible with parameter-efficient adaptation. The same applies to structured modalities, where any mapping method can be used to project features into a higher dimensional space.

\section{Experiments}

We evaluate MAGNUM on several industry-relevant benchmarks collected with the exact purpose of facing real-case scenario challenges. All the benchmark datasets were selected with the only constraint that at least one structured modality needed to be present in conjunction with unstructured modalities.

\subsection{Evaluation Benchmarks} 
A mixture of bimodal and trimodal datasets has been considered as the evaluation benchmark datasets. The datasets are described as follows:
\begin{itemize}
    \item \textbf{Amazon Review - Beauty} \cite{ni2019justifying}. A trimodal vision-language-tabular dataset. It consists of images about commercial products available in the Amazon marketplace in the beauty category, user reviews written in natural language, and a table of meta-data associated with a target user. The dependent variable is the number of review stars left by the target user after purchasing.
    \item \textbf{Amazon Review - Fashion} \cite{ni2019justifying}. A trimodal vision-language-tabular dataset. It consists of images about commercial products available in the Amazon marketplace in the fashion category, user reviews written in natural language, and a table of meta-data associated with a target user. The dependent variable is the number of review stars left by the target user after purchasing.
    \item \textbf{DVM Cars} \cite{huang2022dvm}. A bimodal vision-tabular dataset consisting of images of cars from different angles, and a table of meta-data and descriptive statistics about the given car. The dependent variable is the set of possible car brand.
    \item \textbf{Covid 19} \cite{saltzcovid19}. A bimodal vision-tabular medical dataset that merges patient hospitalization data and demographics with x-ray scan taken during Covid diagnosis. The dependent variable is the clinical status of the patient after hospitalization.
    \item \textbf{Clothings Review} \cite{kagclotrev}. A bimodal language-tabular dataset that consists of user reviews written in natural language associated with meta-data of clothing products. The dependent variable is a binary recommendation for the given product.
    \item \textbf{Hippocorpus} \cite{sap2020recollection}. A bimodal language-tabular psychological dataset containing verbal report written in natural language of memorized and fiction events, together with a table of psychological status scores and demographics of the person reporting the event. The dependent variable is the actual membership of the reported event as whether memorized or fiction.
\end{itemize}

For all the datasets, a 70-15-15 split rule has been used to derive training, validation, and test sets, respectively. The performance evaluation criterion is the balanced accuracy defined as the accuracy score with class-balanced sample weights \cite{brodersen2010balanced}, which is a more appropriate metric for classification tasks with unbalanced classes, as is the case for most of the datasets considered.

\subsection{Implementation Details}

We use the pre trained $16 \times 16$ patches ViT \cite{dosovitskiy2020image} for image processing, and pre-trained RoBERTa \cite{liu2019roberta} for language processing. Both versions are taken from the HuggingFace's Transformers library \cite{wolf-etal-2020-transformers}. Models and pipelines are built in PyTorch. We use the AdamW optimizer, by setting learning rate to $0.00325$, weight decay to $1e^{-5}$, and a cosine annealing with warmup, for all the benchmarks. We set batch size to $8$ and number of epochs to $30$. We pick the checkpoint of the best epoch validation accuracy as the best model to use in inference. Balanced accuracy is then computed on the test set, for every model. All the experiments have been conducted on one single GeForce RTX 3090 Ti.

\subsection{Model comparison}

We compare MAGNUM with a set of competitor models proposed in the literature to solve multimodal tasks in either unstructured-only or unstructured-structured scenarios. We select two representative models with available codes and a production-ready setting: Flava \cite{singh2022flava}, and TaBERT \cite{yin2020tabert}. The former explicitly accepts paired image and text inputs in a dual-encoder architecture and has been pre-trained on a large corpus of vision-language datasets. The latter has been pre-trained on pairs of database data entry structures and database queries expressed in natural language. Tabular inputs have been turned into a sentence in case a model can't natively process table format (e.g. Flava). A third candidate, LANISTR \cite{ebrahimi2023lanistr}, has been excluded since the code is unavailable and no clear instructions were provided to mimic the pre-training pipeline. We compare the performance of the competitor models with MAGNUM on the six benchmark datasets. Results are shown in Table \ref{table:tab1}.

\begin{table*}
    \begin{center}
        \setlength{\tabcolsep}{5pt}
        \renewcommand{\arraystretch}{1.2}
        \begin{tabular}{lcccccc|c}
            \toprule[1.5pt]
            \multirow{2}{*}{\textbf{Model}} & \multicolumn{6}{c|}{\textbf{Dataset}} & \multirow{2}{*}{\textbf{Avg.$\uparrow$}} \\
            \cline{2-7}
            & Amazon Rev B & Amazon Rev F & DVM Cars & Covid 19 & Clothings Rev & Hippocorpus & \\
            \hline
            TaBERT & 0.57 & \textbf{0.60} & 0.20 & 0.72 & 0.92 & 0.67 & 0.61 \\
            Flava & 0.47 & 0.42 & 0.04 & 0.88 & 0.80 & 0.65 & 0.54 \\
            \hline
            MAGNUM & \textbf{0.60} & 0.58 & \textbf{0.72} & \textbf{0.89} & \textbf{0.95} & \textbf{0.80} & \textbf{0.76} \\
            \bottomrule[1.5pt]
        \end{tabular}
    \end{center}
    \caption{Balanced accuracy of the different models on the test sets.}
    \label{table:tab1}
\end{table*}

Results show that MAGNUM outperforms competitor models in the majority of the benchmark datasets by a large margin. A clear advantage of MAGNUM over Flava was expected since Flava has not been tested against datasets compatible with its pre-training scope, i.e. vision-language datasets. However, MAGNUM outperforms TaBERT on its specialized data domain, i.e. tabular-language datasets, such as Clothings Review, and Hippocorpus, and shows comparable performance in one tabular-vision-language benchmark, i.e. Amazon Review Fashion.

\section{Conclusions}

In this paper, we presented MAGNUM, a modular and flexible architecture that natively handles both structured and unstructured data. MAGNUM doesn't require pre-training on multimodal foundational datasets, overcoming the limitation related to the scarcity of structured-unstructured datasets for several use-cases. Furthermore, our model agnostically accepts both pre-trained and non pre-trained architectures. It first leverages parameter-efficient and feature-level projection methods to bring modality-specific inputs to a common dimensional space, and then applies compression and fusion strategies to obtain a final representation of the multimodal input. MAGNUM outperforms multimodal pre-trained architectures in several industry-relevant scenarios. 

{\small
\bibliographystyle{ieee_fullname}
\bibliography{egbib}
}

\end{document}